\documentclass{article} % For LaTeX2e
\usepackage{nips11submit_e,times}
\usepackage{subfigure}
\usepackage{graphicx} \usepackage{amsmath} \usepackage{amssymb}
\usepackage{epic}\usepackage{eepic}
\newcommand{\BEQ}{\begin{equation}}
\newcommand{\EEQ}{\end{equation}}
\newcommand{\BEA}{\begin{eqnarray}}
\newcommand{\EEA}{\end{eqnarray}}
\newcommand{\BGA}{\begin{gather}}
\newcommand{\EGA}{\end{gather}}
\renewcommand{\d}{{\rm d}}
\newcommand{\e}{\epsilon}

\newcommand{\la}{\lambda}

\newcommand{\bx}{{\bf x}}
\newcommand{\by}{{\bf y}}

\newcommand{\comment}[1]{}
\newcommand{\x}{x }
\newcommand{\bs}{{\bf s}}
\newcommand{\X}{{\cal X} }

\renewcommand{\S}{{\cal S} }

\newcommand{\T}{\mathbb{T}}
\newcommand{\kt}{\hat{t}}
\newcommand{\htau}{\hat{\tau}}

\newcommand{\hl}{\hat{\lambda}^n }
\newcommand{\hla}{\hat{\lambda} }
\newcommand{\epf}{\varphi}
\renewcommand{\P}{\mathbb{P}}

\title{Comparative Analysis of Viterbi Training and \\ Maximum Likelihood Estimation for HMMs}
\author{ {\bf Armen Allahverdyan%\thanks{Currently at: {\em Laboratoire de Physique Statistique et Systemes Complexes}, ISMANS, Le Mans, France.} 
} 
\\  
Yerevan Physics Institute \\  
Yerevan, Armenia \\ 
\texttt{aarmen@yerphi.am}
\And 
{\bf Aram Galstyan}  \\ 
USC Information Sciences Institute\\
Marina del Rey, CA, USA \\    
\texttt{galstyan@isi.edu}          
} 
 \nipsfinalcopy
\begin{document} 
 
\maketitle 
 
\begin{abstract} 
We present an asymptotic analysis of Viterbi Training (VT) and contrast
it with a more conventional Maximum Likelihood (ML) approach to
parameter estimation in Hidden Markov Models. While ML estimator works
by (locally) maximizing the likelihood of the observed data, VT seeks to
maximize the probability of the most likely hidden state sequence. We
develop an analytical framework based on a generating function formalism
and illustrate it on an exactly solvable model of HMM with one
unambiguous symbol. For this particular model the ML objective function
is continuously degenerate. VT objective, in contrast, is shown to have
only finite degeneracy. Furthermore, VT converges faster and results in sparser (simpler)
models, thus realizing an automatic Occam's razor for HMM learning.
For more general scenario VT can be worse compared to
ML but still capable of correctly recovering most of the parameters. 
\end{abstract}
 
\section{Introduction}

Hidden Markov Models (HMM) provide one of the simplest examples of structured 
data observed through a noisy channel. The inference problems of HMM naturally divide into two classes
\cite{rabiner_review,ephraim_review}: {\it i)} recovering the hidden
sequence of states given the observed sequence, and {\it ii)} estimating
the model parameters (transition probabilities of the hidden Markov
chain and/or conditional probabilities of observations) from the
observed sequence. The first class of problems is usually solved via the
maximum a posteriori (MAP) method and its computational implementation
known as Viterbi algorithm \cite{rabiner_review,ephraim_review}. For the
parameter estimation problem, the prevailing method is maximum
likelihood (ML) estimation, which finds the parameters  by maximizing the
likelihood of the observed data. Since global optimization is generally
intractable, in practice it is implemented through an
expectation--maximization (EM) procedure known as Baum--Welch
algorithm~\cite{rabiner_review,ephraim_review}. 

An alternative approach to parameter learning is Viterbi Training (VT),
also known in the literature as segmental K-means, Baum--Viterbi
algorithm, classification EM, hard EM, etc. Instead of maximizing the likelihood of the observed data, VT seeks to maximize the probability of the most likely hidden state
sequence. Maximizing VT objective function is hard~\cite{Cohen2010}, so in practice it  is implemented via an EM-style iterations between calculating the MAP sequence and adjusting the model parameters based on the sequence statistics. It is known that VT lacks some of the desired features of ML estimation such as consistency~\cite{leroux}, and in fact, can produce biased estimates~\cite{ephraim_review}. However, it has been shown to
perform well in practice, which explains its widespread use in
applications such as speech recognition~\cite{juang}, unsupervised dependency parsing~\cite{Spitkovsky2010}, grammar
induction~\cite{Benedi2005}, ion channel modeling~\cite{Qin2004}. It is
generally assumed that VT is more robust and faster but usually less
accurate, although for certain tasks it outperforms conventional
EM~\cite{Spitkovsky2010}. 

The current understanding of when and under what circumstances one method should be preferred over the other is not well--established. For HMMs with continuos observations, Ref.~\cite{Merhav1991} established an upper bound on the difference between the ML and VT objective functions, and showed that both approaches produce asymptotically  similar  estimates when the dimensionality of the observation space is very large. Note, however, that this asymptotic limit is not very interesting as it  makes the structure imposed by the Markovian process irrelevant. A similar attempt to compare both approaches on discrete models (for stochastic context free grammars) was  presented in~\cite{Sanchez1996}. However, the established bound was very loose.

Our goal here is to understand, both qualitatively and quantitatively, the difference between the two estimation methods. We  develop an analytical approach based on generating functions for examining the asymptotic properties of both approaches. Previously, a similar approach  was used  for calculating entropy rate of a hidden Markov process~\cite{armen}. Here we provide a non-trivial extension of the methods that allows to perform comparative asymptotic analysis of ML and VT estimation.  It is shown that  both estimation methods correspond to certain free-energy minimization problem at different {\em temperatures}.  Furthermore, we demonstrate the approach on a particular class of HMM with one unambiguous symbol and obtain a closed--form solution to the estimation problem. This class  of HMMs is sufficiently rich so as to include models where not all parameters can be determined from the observations, i.e., the model is not {\em identifiable}~\cite{1957,ito,ephraim_review}.

We find that for the considered model VT  is a better option if the 
ML objective  is degenerate (i.e., not all parameters can be obtained
from observations). Namely, not only VT recovers the identifiable parameters but it also provides a simple (in the sense that non-identifiable
parameters are set to zero) and optimal (in the sense of the MAP
performance) solution. Hence, VT  realizes an automatic Occam's razor for the HMM 
learning. In addition, we show that the VT algorithm  for this model converges faster than the conventional EM approach. Whenever the ML objective is not degenerate, VT  leads generally to inferior results that, nevertheless, may be partially correct in the sense of recovering certain (not all) parameters.

\section{Hidden Markov Process}
\label{def}
Let $\S=\{\S_0,\S_1,\S_2,...\}$ be a discrete-time, stationary, Markov  process 
with conditional probability 
\BEA
{\rm Pr}[\S_{k+l}=s_{k}|\S_{k-1+l}=s_{k-1}]
=p(s_{k}|s_{k-1}),
\label{arus}
\EEA
where $l$ is an integer. Each realization $s_k$ of the
random variable $\S_k$ takes values $1,...,L$. 
We assume that $\S$ is mixing:
it has a unique stationary distribution $p_{\rm st}(s)$,
${\sum}_{r=1}^L p(s|r) p_{\rm st }(r)=p_{\rm st }(s)$,
that is established from any initial probability in the long time limit.

Let random variables $\X_i$, with realizations $\x_i=1,..,M$, be noisy
observations of $\S_i$: the (time-invariant) conditional probability of
observing $\X_i=\x_i$ given the realization $\S_i=s_i$ of the Markov
process is $\pi(\x_k|s_k)$. Defining $\bx \equiv (\x_N,...,\x_1)$, $\bs \equiv (s_N,...,s_0)$,
the joint probability of $\S,\X$ reads
\BEA
\label{1}
P({\bf s}, {\bf x})
= T_{s_{N}\,s_{N-1}}(\x_{N})...
T_{s_{1}\,s_{0}}(\x_1)\, p_{\rm st}(s_0), 
\EEA
where the $L\times L$ transfer-matrix $T(x)$ with matrix elements 
$T_{s_{i}\,s_{i-1}}(\x)$ is defined as
\BEA
\label{transfer}
T_{s_{i}\,s_{i-1}}(\x)= \pi(\x|s_i)\, p(s_{i}|s_{i-1}).
\EEA
$\X=\{\X_1,\X_2,...\}$ is called a hidden Markov process. Generally, it is not Markov, but 
it inherits stationarity and mixing from $\S$ \cite{ephraim_review}. 
The probabilities for $\X$ can be represented as follows: 
\BEA
\label{mubarak}
\label{2}
P(\bx)={\sum}_{ss'}\left[\T(\bx) \right]_{ss'} p_{\rm st}(s'), ~~~
\T(\bx)\equiv T(\x_N)T(\x_{N-1})\ldots T(\x_1),
\EEA
where $\T(\bx)$ is a product of transfer matrices.

\section{Parameter Estimation}
\label{para}

\subsection{Maximum Likelihood Estimation}
The unknown parameters of an HMM are the transition probabilities $p(s|s')$ of the
Markov process and the observation probabilities $\pi(x|s)$; see
(\ref{1}). They have to be estimated from
the observed sequence $\bx$. This is standardly done via
the maximum-likelihood approach: one starts with some trial
values $\hat{p}(s|s')$, $\hat{\pi}(x|s)$ of the parameters and
calculates the (log)-likelihood $\ln \hat{P}(\bx)$, where $\hat{P}$ means
the probability (\ref{2}) calculated at the trial values of the
parameters. Next, one maximizes $\ln \hat{P}(\bx)$ over $\hat{p}(s|s')$
and $\hat{\pi}(x|s)$ for the given observed sequence $\bx$ (in practice 
this is done via the Baum-Welch algorithm \cite{rabiner_review,ephraim_review}). 
The rationale of this approach is as follows. Provided that the length
$N$ of the observed sequence is long, and recaling that $\X$ is mixing 
(due to the analogous feature of $\S$) we get probability-one convergence (law of large numbers)
\cite{ephraim_review}:
\BEA
\label{vse}
\ln \hat{P}(\bx)\to {\sum}_{\by} P(\by)\ln \hat{P}(\by),
\EEA
where the average is taken over the true probability $P(...)$ that generated $\bx$. Since the relative entropy is non-negative,
${\sum}_{\bx} P(\bx)\ln [{P(\bx)}/{\hat{P}(\bx)}]\geq 0$,
the global maximum of ${\sum}_{\bx} P(\bx)\ln \hat{P}(\bx)$ as a
function of $\hat{p}(s|s')$ and $\hat{\pi}(x|s)$ is reached for
$\hat{p}(s|s')=p(s|s')$ and $\hat{\pi}(x|s)=\pi(x|s)$. This argument is 
silent on how unique this global maximum is
and how difficult to reach it. 

\subsection{Viterbi Training}
\label{vitus}
An alternative approach to the parameter learning employs the maximal a posteriori
(MAP) estimation and proceeds as follows: Instead of maximizing the likelihood of
observed data~(\ref{vse}) one tries to maximize the probability of the
most likely sequence~\cite{rabiner_review,ephraim_review}. Given the
joint probability $\hat{P}({\bf s}, {\bf x})$ at trial values of
parameters, and given the observed sequence ${\bf x}$, one estimates the
generating state-sequence ${\bf s}$ via maximizing the a posteriori
probability
\BEA
\hat{P}({\bf s}| {\bf x}) ={\hat{P}({\bf s}, {\bf x}) }/{\hat{P}({\bf x}) }
\EEA
over ${\bf s}$. Since $\hat{P}({\bf x})$ does not depend on ${\bf s}$, one can maximize $\ln\hat{P}({\bf s}, {\bf x}) $.
If the number of observations is sufficiently large $N\to\infty$, one can substitute 
${\rm max}_{\bf s}\ln\hat{P}({\bf s}, {\bf x})$ by its average over $P(...)$ [see (\ref{vse})] 
and instead maximize (over model parameters)
\BEA
{\sum}_{\bx}P(\bx)\,{\rm max}_{{\bf s}}\ln \hat{P} ({\bf s}, {\bf x}) .
\label{ko2}
\EEA
To relate (\ref{ko2}) to the free energy concept {(see e.g. \cite{aa,dna}), we define an auxiliary (Gibbsian) probability 
\BEA
\label{lamarc}
\hat{\rho}_\beta ({\bf s}|{\bf x})= {\hat{P}^\beta({\bf s}, {\bf x})}/\left[{{\sum}_{\bf s'} \hat{P}^\beta({\bf s'}, {\bf x})   }\right],
\EEA
where $\beta>0$ is a parameter. As a function of $\bs$ (and for a fixed $\bx$), $\hat{\rho}_{\beta\to\infty} ({\bf s}|{\bf x})$
concentrates on those $\bs$ that maximize $\ln\hat{P}({\bf s}, {\bf x}) $:
\BEA
\label{darwin}
\hat{\rho}_{\beta\to\infty}({\bf s}|{\bf x})\to \frac{1}{\cal N}{\sum}_j \delta[{\bf s},\,\widetilde{{\bf s}}^{[j]}({\bf x})],
\EEA
where $\delta(s,s')$ is the Kronecker delta, $\widetilde{{\bf s}}^{[j]}({\bf x})$ 
are equivalent outcomes of the maximization, and ${\cal N}$ is the number of such outcomes.
Further, define
\BEA
F_\beta\equiv -
\frac{1}{\beta}{\sum}_{\bx}P(\bx)\, \ln {\sum}_{\bs} \hat{P}^\beta({\bs}, {\bx}) .
\label{gnorr}
\EEA
Within statistical mechanics Eqs.~\ref{lamarc} and~\ref{gnorr} refer to,
respectively, the {\em Gibbs distribution} and {\em free energy} of a
physical system with Hamiltonian $H=-\ln P({\bf s}, {\bf x})$ coupled to
a thermal bath at inverse temperature $\beta=1/T$ \cite{aa,dna}. It is then
clear that ML and Viterbi Training correspond to minimizing the free
energy Eq.~\ref{gnorr} at $\beta=1$ and $\beta=\infty$, respectively.
Note that $\beta^2\partial_\beta F=-{\sum}_{\bf x}P(\bf x){\sum}_{\bf s}
\rho_\beta ({\bf s}|x)\ln \rho_\beta ({\bf s}|x) \geq 0$, which yields
$F_1\leq F_\infty$. 

\subsection{Local Optimization}

As we mentioned, global maximization of neither objective is feasible in
the general case. Instead, in practice this maximization is {\it
locally} implemented via an EM-type algorithm
\cite{rabiner_review,ephraim_review}: for a given observed sequence
${\bf x}$, and for some initial values of the parameters, one calculates
the expected value of the objective function under the trial parameters
(E-step), and adjusts the parameters to maximize this expectation
(M-step). The resulting estimates of the parameters are now employed as
new trial parameters and the previous step is repeated. This recursion
continues till convergence. 

For our purposes,  this procedure can be understood as calculating certain statistics of the hidden sequence averaged over the Gibbs distribution Eqs.~\ref{lamarc}. Indeed, let us introduce
$f_\gamma(\bs)\equiv e^{\beta\gamma{\sum}_{i=1}^N \delta(s_{i+1},a) \delta(s_i,b)  }$ and define
\BEA
\beta F_\beta(\gamma)\equiv -
{\sum}_{\bx}P(\bx)\, \ln {\sum}_{\bs} \hat{P}^\beta(\bs, \bx)f_\gamma(\bs).
\label{mozart}
\EEA
Then, for instance, the (iterative) Viterbi estimate of the transition probabilities  are given as follows:
\BEA
\widetilde{P}({\cal S}_{k+1}=a,\, {\cal S}_{k}=b)=
-\partial_\gamma [F_\infty(\gamma)]|_{\gamma\to 0}.
\label{salieri}
\EEA
Conditional probabilities for observations are calculated similarly via a different indicator function.

\section{Generating Function }
\label{tro}

Note from (\ref{mubarak}) that both $P(\bx)$ and $\hat{P}(\bx)$ are
obtained as matrix-products. For a large number of multipliers the
behavior of such products is governed by the multiplicative law of large
numbers. We now recall its formulation from \cite{goldsheid}: for $N\to
\infty$ and $\bx$ generated by the mixing process $\X$ there is a
probability-one convergence:
\BEA
\label{lq} \frac{1}{N}\ln  || \T(\bx) || \to 
\frac{1}{N}{\sum}_\by P(\by) \ln\la[\T(\by)],
\label{dag}
\EEA
where $||...||$ is a matrix norm in the linear space of $L\times L$
matrices, and $\la[\T(\bx)]$ is the maximal eigenvalue of $\T(\bx)$.
Note that (\ref{dag}) does not depend on the specific norm chosen,
because all norms in the finite-dimensional linear space are equivalent;
they differ by a multiplicative factor that disappears for
$N\to\infty$ \cite{goldsheid}. Eqs.~(\ref{mubarak}, \ref{dag}) also imply ${\sum}_{\bx} \la[\T(\bx)]\to 1$. Altogether,
we calculate (\ref{vse}) via its probability-one limit
\BEA
\label{pri}
\frac{1}{N}{\sum}_{\bx} P(\bx)\ln \hat{P}(\bx)\to
\frac{1}{N}{\sum}_{\bx} \la[\T(\bx)]\ln \la[\hat{\T}(\bx)].
\EEA
Note that the multiplicative law of large numbers is normally formulated
for the maximal singular value. Its reformulation in terms of the maximal
eigenvalue needs additional arguments \cite{armen}.

Introducing the generating function
\BEA
\label{4}
\Lambda^N(n,N)
={\sum}_{\bx} \lambda [ \T(\bx) ]\lambda^n \left[ \hat{\T}(\bx) \right],
\EEA
where $n$ is a non-negative number, and where $\Lambda^N(n,N)$ means $\Lambda(n,N)$
in degree of $N$, one represents (\ref{pri}) as
\BEA
\label{kepler}
\frac{1}{N}{\sum}_{\bx} \la[\T(\bx)]\ln \la[\hat{\T}(\bx)]=
\partial_{n} \Lambda(n,N) |_{n=0},
\EEA
where we took into account $\Lambda(0,N)=1$, as follows from (\ref{4}).

The behavior of $\Lambda^N(n,N)$ is better understood after 
expressing it via the zeta-function $\xi(z,n)$ \cite{armen}
\BEA
\label{zeta_meta}
\xi(z,n)=\exp\left[-
{\sum}_{m=1}^\infty \frac{z^m}{m}\Lambda^m(n,m)
\right],
\EEA
where $\Lambda^m(n,m)\geq 0$ is given by (\ref{4}). 
Since for a large $N$, $\Lambda^N(n,N)\to \Lambda^N(n)$ [this is the content of (\ref{lq})], the zeta-function
$\xi(z,n)$ has a zero at $z=\frac{1}{ \Lambda(n) }$: 
\BEA
\xi({1}/{\Lambda(n)},\, n)=0.
\label{khrych}
\EEA
Indeed for $z$ close (but smaller than) $\frac{1}{ \Lambda(n) }$, the series
${\sum}_{m=1}^\infty \frac{z^m}{m}\Lambda^m(n,m)\to 
{\sum}_{m=1}^\infty \frac{[z\Lambda(n)]^m}{m}$ almost diverges and
one has $\xi(z,n)\to 1-z\Lambda(n)$.
Recalling that $\Lambda(0)=1$ and taking $n\to 0$ in $0=\frac{\d}{\d n}\xi (\frac{1}{\Lambda(n)},n)$,
we get from (\ref{kepler})
\BEA
\label{mkno}
\frac{1}{N}{\sum}_{\bx} \la[\T(\bx)]\ln \la[\hat{\T}(\bx)]=
\frac{{\partial_n}\xi(1,0)}{{\partial_z}\xi(1,0)}.
\EEA
For calculating $-\beta F_\beta$ in (\ref{gnorr}) we have
instead of (\ref{mkno})
\BEA
\label{mkno1}
-\frac{\beta F_\beta}{N}=
\frac{{\partial_n}\xi^{[\beta]}(1,0)}{{\partial_z}\xi^{[\beta]}(1,0)},
\EEA
where $\xi^{[\beta]}(z,n)$ employs
$\hat{T}^{\beta}_{s_{i}\,s_{i-1}}(\x)= \hat{\pi}^\beta(\x|s_i)\, \hat{p}^\beta(s_{i}|s_{i-1})$
instead of $\hat{T}_{s_{i}\,s_{i-1}}(\x)$ in (\ref{mkno}).

Though in this paper we restricted ourselves to the limit $N\to\infty$,
we stress that the knowledge of the generating function $\Lambda^N(n,N)$
allows to analyze the learning algorithms for any finite
$N$.

\section{Hidden Markov Model with One Unambiguous Symbol}
\label{model}
\subsection{Definition}
\label{garun_al_rashid}
\begin{figure}%[hb]
%\vspace{0.2cm}
\center
\includegraphics[width=5cm]{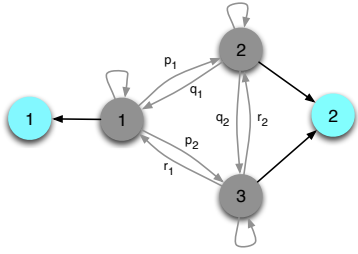}
\caption{{\it 
The hidden Markov process
(\ref{mack1}--\ref{mack2}) for $\e=0$. Gray circles and arrows indicate
on the realization and transitions of the internal Markov process; see (\ref{mack1}). 
The light circles and black arrows indicate on the realizations of the observed process. 
}}
\hfill
\label{f1}
\end{figure}

Given a $L$-state Markov process $\S$, the observed
process $\X$ has two states $1$ and $2$; see Fig.~1. All internal states besides
one are observed as $2$, while the internal state $1$ produces, respectively, $1$ and
$2$ with probabilities $1-\epsilon$ and $\epsilon$. For $L=3$ we obtain from (\ref{arus})
$\pi(1|1)=1-\pi(2|1)=1-\epsilon$, $\pi(1|2)=\pi(1|3)=\pi(2|1)=0$,
$\pi(2|2)=\pi(2|3)=1$. Hence $1$ 
is unambiguous: if it is observed, 
the unobserved process $\S$ was certainly in $1$; see Fig.~1. 
The simplest example of such HMM exists already for $L=2$; see~\cite{han_marcus} for analytical features of entropy for
this case. We, however, describe in detail the $L=3$ situation, since this case will be seen to be generic
(in contrast to $L=2$) and it allows straightforward generalizations to $L>3$. 
The transition matrix (\ref{arus}) of a general $L=3$ Markov process reads
\begin{gather}
\label{mack1}
\P\equiv\{\, p(s|s')\, \}_{s,s'=1}^3=
\left(\begin{array}{rrr}
p_0 & q_1 & r_1 \\
p_1 & q_0  & r_2 \\
p_2 & q_2   & r_0  \\
\end{array}\right), ~~~
\left(\begin{array}{r}
p_0 \\
q_0 \\
r_0  \\
\end{array}\right)=
\left(\begin{array}{r}
1-p_1-p_2 \\
1-r_1-r_2 \\
1-r_1-r_2  \\
\end{array}\right)
\end{gather}
where, e.g., $q_1=p(1|2)$ is the transition probability $2\to 1$; see Fig.~1.
The corresponding transfer matrices read from (\ref{transfer})
\BEA
\label{mack2}
T(1)=(1-\e) 
\left(\begin{array}{rrr}
p_0 & q_1 & r_1 \\
0   & 0   & 0  \\
0   & 0   & 0   \\
\end{array}\right), ~~~ T(2)=\P-T(1).
\EEA
Eq.~(\ref{mack2}) makes straightforward the reconstruction of the
transfer-matrices for $L\geq 4$. It should also be obvious that for all $L$ only the first row
of $T(1)$ consists of non-zero elements. 

For $\epsilon=0$ we get from (\ref{mack2}) the simplest example of an
aggregated HMM, where several Markov states are mapped
into one observed state. This model plays a special role for the HMM
theory, since it was employed in the pioneering study of the
non-identifiability problem \cite{1957}. 

\subsection{Solution of the Model}
For this model $\xi(z,n)$ can be calculated exactly, because $T(1)$ has only one non-zero row.
Using the method outlined in the supplementary material (see also \cite{armen,artuso}) we get
\BEA
\label{raa}
\xi(z,n)=1-z(  t_0 \kt^n_0+ \tau_0 \htau^n_0  ) +{\sum}_{k=2}^\infty [\tau \htau^n \kt^n_{k-2}t_{k-2}-\kt^n_{k-1} t_{k-1}   ] z^k
\EEA
where $\tau$ and $\htau$ are the largest eigenvalues of $T(2)$ and $\hat{T}(2)$, respectively 
\begin{gather}
\label{mo}
t_k\equiv\langle 1 | T(1)T(2)^{k}| 1 \rangle = {\sum}_{\alpha=1}^L \tau_\alpha^k \psi_\alpha, \\
\psi_\alpha \equiv \langle 1 | T(1)|R_\alpha\rangle\langle L_\alpha|1\rangle, ~~~~~ \langle 1|\equiv (1,0,\ldots,0).
\label{momo}
\end{gather}
Here $|R_\alpha\rangle$ and $\langle L_\alpha|$ are, respectively right
and left eigenvalues of $T(2)$, while $\tau_1,\ldots,\tau_L$ ($\tau_L\equiv\tau$) are the
eigenvalues of $T(2)$. Eqs.~(\ref{mo}, \ref{momo}) obviously extend to hatted quantities.

We get from (\ref{raa}, \ref{mkno}):
\begin{gather}
\xi(1,n) = (1-\hat{\tau}^n\tau) \left(1-{\sum}_{k=0}^\infty \kt_k^n t_k \right),\\
\frac{\partial_n\xi(1,0)}{\partial_z\xi(1,0)}=\frac{{\sum}_{k=0}^\infty t_k\ln [\hat{t}_k]}{{\sum}_{k=0}^\infty(k+1)t_k}.
\label{rabinovich}
\end{gather}
Note that for $\epsilon=0$, $t_k$ are return probabilities to the state
$1$ of the $L$-state Markov process. For $\epsilon>0$ this 
interpretation does not hold, but $t_k$ still has a meaning of
probability as ${\sum}_{k=0}^\infty t_k=1$. 

Turning to equations (\ref{mkno}, \ref{rabinovich}) for the free energy, 
we note that as a function of trial values it depends on the following $2L$ parameters:
\BEA
(\hat{\tau}_1,\ldots,\hat{\tau}_L,\hat{\psi}_1,\ldots,\hat{\psi}_L).
\label{me}
\EEA
As a function of the true values, the free energy depends on the same
$2L$ parameters (\ref{me}) [without hats], though concrete
dependencies are different. For the studied class of HMM there are at
most $L(L-1)+1$ unknown parameters: $L(L-1)$ transition
probabilities of the unobserved Markov chain, and one parameter
$\e$ coming from observations. We checked numerically
that the Jacobian of the transformation from the unknown
parameters to the parameters (\ref{me}) has rank $2L-1$. 
Any $2L-1$ parameters among (\ref{me}) can be taken as independent ones. 

For $L>2$ the number of {\it effective} independent parameters that affect the free energy  is smaller than the number of
 parameters. So if the number of unknown parameters is larger
than $2L-1$, neither of them can be found explicitly. One can only
determine the values of the effective parameters. 

\section{The Simplest Non-Trivial Scenario}
\label{tri}
The following example allows the full analytical treatment, 
but is generic in the sense that it contains all the 
key features of the more general situation given above (\ref{mack1}).
Assume that $L=3$ and
\BEA
\label{truism}
q_0=\hat{q}_0=r_0=\hat{r}_0=0, \qquad \e=\hat{\e}=0.
\EEA
Note the following explicit expressions
\BEA
\label{kappa}
t_0 =p_0, 
~
t_1=p_1q_1+p_2r_1, 
~
t_2 = p_1r_1q_2+p_2q_1r_2, \\
\label{kappa1}
\tau=\tau_3=\sqrt{q_2r_2}, ~~~ \tau_2=-\tau, \quad \tau_1=0,\\
\psi_3  - \psi_2 = {t_1}/{\tau}, ~~~ \psi_3  + \psi_2 = {t_2}/{\tau^2},
\label{kappa2}
\EEA
Eqs.~(\ref{kappa}--\ref{kappa2}) with obvious changes $s_i\to\hat{s}_i$ 
for every symbol $s_i$ hold for $\kt_k$, $\htau_k$ and $\hat{\psi}_k$. 
Note a consequence of
${\sum}_{k=0}^2p_k={\sum}_{k=0}^2q_k={\sum}_{k=0}^2r_k=1$:
\BEA 
\tau^2(1-t_0)=1-t_0-t_1-t_2.
\label{katar}
\EEA
\subsection{Optimization of $F_1$}
Eqs.~(\ref{rabinovich}) and (\ref{kappa}--\ref{kappa2}) imply ${\sum}_{k=0}^\infty (k+1) t_k=\frac{\mu}{1-\tau^2}$, 
\BEA
&&\mu\equiv 1-\tau^2+t_2+(1-t_0)(1+\tau^2)>0,\\
\label{xosrov}
&&-\frac{\mu F_1}{N}
=t_1\ln\kt_1+t_2\ln\kt_2+(1-\tau^2)t_0\ln\kt_0+(1-t_0)\tau^2\ln\htau^2 \ .
\EEA
The free energy  $F_1$ depends on three independent parameters
$\kt_0,\kt_1,\kt_2$ [recall (\ref{katar})]. Hence, minimizing $F_1$ we
get $\kt_i=t_i$ ($i=0,1,2$), but we do not obtain a definite solution for the
unknown parameters: any four numbers $\hat{p}_1$, $\hat{p}_2$,
$\hat{q}_1$, $\hat{r}_1$ satisfying three equations
$t_0 =1-\hat{p}_1-\hat{p}_2$, $t_1=\hat{p}_1\hat{q}_1+\hat{p}_2\hat{r}_1$, 
$t_2 = \hat{p}_1\hat{r}_1(1-\hat{q}_1)+\hat{p}_2\hat{q}_1(1-\hat{r}_1)$,
minimize $F_1$. 

\subsection{Optimization of $F_\infty$}
In deriving (\ref{xosrov}) we used no particular feature of
$\{\hat{p}_k\}_{k=0}^2$, $\{\hat{q}_k\}_{k=1}^2$,
$\{\hat{r}_k\}_{k=1}^2$. Hence, as seen from
(\ref{mkno1}), the free energy at $\beta>0$ is recovered from
(\ref{xosrov}) by equating its LHS to $-\frac{\beta F_\beta}{N}$ and by
taking in its RHS: 
$\kt_0 \to \hat{p}_0^\beta$,  $\htau^2\to \hat{q}_2^\beta\hat{r}_2^\beta$,
$\kt_1\to\hat{p}_1^\beta \hat{q}_1^\beta+\hat{p}_2^\beta \hat{r}_1^\beta$, 
$\kt_2 \to \hat{p}_1^\beta\hat{r}_1^\beta\hat{q}_2^\beta+\hat{p}_2^\beta\hat{q}_1^\beta\hat{r}_2^\beta$.
The zero-temperature free energy reads from (\ref{xosrov})
\BEA
-\frac{\mu F_\infty}{N}
=(1-\tau^2)t_0\ln\kt_0+(1-t_0)\tau^2\ln\htau^2 &+&t_1\ln{\rm max}[\hat{p}_1\hat{q}_1,\hat{p}_2\hat{r}_1] \nonumber \\
&+&t_2\ln{\rm max}[\hat{p}_2\hat{q}_1\hat{r}_2,\hat{p}_1\hat{r}_1\hat{q}_2].
\label{akan}
\EEA
We now 
minimize $F_\infty$ over the trial parameters 
$\hat{p}_1,\hat{p}_2,\hat{q}_1,\hat{r}_1$. This is not 
what is done by the VT algorithm; see the 
discussion after (\ref{salieri}). But at any rate both
procedures aim to minimize the same target. VT recursion  for this models will be studied in section \ref{bmw} | it leads to the same result.
Minimizing $F_\infty$ over the trial parameters 
produces four distinct solutions:
\BEA
\{\hat{\sigma}_i\}_{i=1}^4=\{\hat{p}_1=0, \, \hat{p}_2=0, \, \hat{q}_1=0, \, \hat{r}_1=0\}.
\label{haso}
\EEA
For each of these four solutions: $\hat{t}_i=t_i$ ($i=0,1,2$) and
$F_1=F_\infty$. The easiest way to get these results is to minimize
$F_\infty$ under conditions $\hat{t}_i=t_i$ (for $i=0,1,2$), obtain
$F_1=F_\infty$ and then to conclude that due to the inequality $F_1 \le F_\infty $ the
conditional minimization led to the global minimization. The logics of
(\ref{haso}) is that the unambiguous state tends to get detached from
the ambiguous ones, since the probabilities nullifying in (\ref{haso})
refer to transitions from or to the unambiguous state. 
\comment{The statistical mechanical meaning of nullifications in (\ref{haso}) is that at zero temperature certain transitions {\it freeze}, much akin to the freezing of molecular motion in  low-temperature substances.}

Note that although minimizing  either $F_\infty$ and $F_1$ produces  correct values
of the independent variables $t_0$, $t_1$, $t_2$, in the
present situation minimizing $F_\infty$ is preferable, because it leads
to the four-fold degenerate set of solutions (\ref{haso}) instead of the
continuously degenerate set.  For instance, if the
solution with $\hat{p}_1=0$ is chosen we get for other parameters
\BEA
\label{gulpa}
\hat{p}_2=1-t_0, ~~ \hat{q}_1= \frac{t_2}{1-t_0-t_1}, ~~ \hat{r}_1= \frac{t_1}{1-t_0}.
\EEA
Furthermore, a more elaborate analysis reveals that for each fixed set of
correct parameters only one among the four solutions Eq.~\ref{haso} provides the
best value for the quality of the MAP reconstruction, i.e. for the
overlap between the original and MAP-decoded sequences.

Finally, we note that minimizing $F_\infty$ allows one to get the correct values $t_0,
t_1, t_2$ of the independent variables $\hat{t}_0$, $\hat{t}_1$ and
$\hat{t}_2$ only if their number is less than the number of unknown
parameters. This is not a drawback, since once the number of
unknown parameters is sufficiently small [less than four for the
present case (\ref{truism})] their exact values are obtained by
minimizing $F_1$. Even then, the minimization of $F_\infty$ can
provide partially correct answers. Assume in (\ref{akan}) that the
parameter $\hat{r}_1$ is known, $\hat{r}_1=r_1$. Now $F_\infty$ has
three local minima  given by $\hat{p}_1=0$, $\hat{p}_2=0$ and
$\hat{q}_1=0$; cf. with (\ref{haso}). The minimum with $\hat{p}_2=0$ is
the global one and it allows to obtain the exact values of the two
effective parameters: $\hat{t}_0=1-\hat{p}_1=t_0$ and
$\hat{t}_1=\hat{p}_1\hat{q}_1=t_1$.  These effective parameters are recovered,
because they do not depend on the known parameter $\hat{r}_1=r_1$.
Two other minima have greater values of $F_\infty$, and they allow to
recover only one effective parameter: $\hat{t}_0=1-\hat{p}_1=t_0$. If in addition
to $\hat{r}_1$ also $\hat{q}_1$ is known, the two local minimia of
$F_\infty$ ($\hat{p}_1=0$ and $\hat{p}_2=0$) allow to recover $\hat{t}_0=t_0$ only. In contrast, if
$\hat{p}_1=p_1$ (or $\hat{p}_2=p_2$) is known exactly, there are three
local minima again|$\hat{p}_2=0$, $\hat{q}_1=0$, $\hat{r}_1=0$|but now
none of effective parameters is equal to its true value: $\hat{t}_i\not= t_i$
($i=0,1,2$).

\subsection{Viterbi EM}
\label{bmw}
Recall the description of the VT algorithm given after
(\ref{salieri}).  For calculating $\widetilde{P}({\cal S}_{k+1}=a,\,
{\cal S}_{k}=b)$ via (\ref{mozart}, \ref{salieri}) we modify the
transfer matrix element in (\ref{4}, \ref{zeta_meta}) as $\hat{T}_{ab}(k)\to
\hat{T}_{ab}(k)e^{\gamma}$, which produces from (\ref{mozart}, \ref{salieri}) 
for the MAP-estimates of the transition probabilities
\begin{gather}
\label{tomas1}
\widetilde{p}_1= \frac{t_1\hat{\chi}_1+t_2\hat{\chi}_2}{t_1+t_2+t_0(1-\tau^2)}, ~~~ \widetilde{p}_2=1-t_0-\widetilde{p}_1,\\
\widetilde{q}_1= \frac{t_1\hat{\chi}_1+t_2(1-\hat{\chi}_2)}{t_1\hat{\chi}_1+t_2+(1-t_0)\tau^2},~~~ \widetilde{q}_2=1-\widetilde{q}_1 \\
\widetilde{r}_1= \frac{t_1(1-\hat{\chi}_1)+t_2\hat{\chi}_2}{t_2+t_1(1-\hat{\chi}_1)+(1-t_0)\tau^2} ~~~ \widetilde{r}_2=1-\widetilde{r}_1, 
\label{tomas2}
\end{gather}
where 
$\hat{\chi}_1 \equiv\frac{\hat{p}_1^\beta\hat{q}_1^\beta}{\hat{p}_1^\beta\hat{q}_1^\beta+\hat{p}_2^\beta\hat{r}_1^\beta}$,
$\hat{\chi}_2 \equiv\frac{\hat{p}_1^\beta\hat{r}_1^\beta\hat{q}_2^\beta}{\hat{p}_1^\beta\hat{r}_1^\beta\hat{q}_2^\beta
+\hat{p}_2^\beta\hat{r}_2^\beta\hat{q}_1^\beta}$.
The $\beta\to\infty$ limit of $\hat{\chi}_1$ and $\hat{\chi}_2$ is
obvious: each of them is equal to $0$ or $1$ depending on the ratios
$\frac{\hat{p}_1\hat{q}_1}{\hat{p}_2\hat{r}_1}$ and $\frac{ \hat{p}_1
\hat{r}_1 \hat{q}_2 }{ \hat{p}_2 \hat{r}_2 \hat{q}_1}$.  
The EM approach amounts to starting with some trial values
$\hat{p}_1$, $\hat{p}_2$, $\hat{q}_1$, $\hat{r}_1$ and using
$\widetilde{p}_1$, $\widetilde{p}_2$, $\widetilde{q}_1$,
$\widetilde{r}_1$ as new trial parameters (and so on). We see
from (\ref{tomas1}--\ref{tomas2}) that the algorithm converges just in
one step: (\ref{tomas1}--\ref{tomas2}) are equal to the parameters given
by one of four solutions (\ref{haso})|which one among the solutions
(\ref{haso}) is selected depends on the on initial trial
parameters in (\ref{tomas1}--\ref{tomas2})|recovering the correct
effective parameters (\ref{kappa}--\ref{kappa2}); e.g.  cf.
(\ref{gulpa}) with (\ref{tomas1}, \ref{tomas2}) under
$\hat{\chi}_1=\hat{\chi}_2=0$. Hence, VT converges in one step in contrast
to the Baum-Welch algorithm (that uses EM to locally minimize $F_1$) which, for the present model, obviously does 
not converge in one step. 
There is possibly a deeper point in the one-step convergence that can
explain why in practice VT converges faster than the Baum-Welch
algorithm \cite{ephraim_review,torres}: recall that, e.g. the Newton
method for local optimization works precisely in one step for quadratic
functions, but generally there is a class of functions, where it
performs faster than (say) the steepest descent method. Further research
should show whether our situation is similar: the VT works just in one
step for this exactly solvable HMM model that belongs to a class of
models, where VT generally performs faster than ML.

We conclude this section by noting that the solvable case (\ref{truism}) is generic: 
its key results extend to the general situation defined
above (\ref{mack1}). We checked this fact numerically for several values
of $L$. In particular, the minimization of $F_\infty$ nullifies as many
trial parameters as necessary to express the remaining parameters via
independent effective parameters $t_0,t_1,\ldots$. Hence for $L=3$ and
$\epsilon=0$ two such trial parameters are nullified; cf. with
discussion around (\ref{me}). If the true  error probability
$\epsilon\not =0$, the trial value $\hat{\epsilon}$ is among the
nullified parameters. Again, there is a discrete degeneracy in solutions
provided by minimizing $F_\infty$. 
\section{Summary}
\label{sum}
We presented a method for analyzing two basic techniques for
parameter estimation in HMMs, and illustrated it on a specific class of
HMMs with one unambiguous symbol. The virtue of this class of
models is that it is exactly solvable, hence the sought quantities can
be obtained in a closed form via generating functions. This is a rare
occasion, because characteristics of HMM such as likelihood or entropy
are notoriously difficult to calculate explicitly \cite{armen}. An important feature of the example considered here is that the set of
unknown parameters is not completely identifiable in the maximum
likelihood sense~\cite{1957,ito}. This corresponds to the zero
eigenvalue of the Hessian for the ML (maximum-likelihood) objective
function. In practice, one can have weaker degeneracy of the objective
function resulting in very small values for the Hessian eigenvalues.
This scenario occurs  {\it often} in various models of physics and computational
biology~\cite{sethna}.  Hence, it is a drawback that the theory of HMM
learning was developed assuming complete identifiably \cite{baum}.

One of our main result is that in contrast to the ML approach that
produces continuously degenerate solutions, VT results in finitely
degenerate solution that is sparse, i.e., some [non-identifiable]
parameters are set to zero, and, furthermore, converges faster. Note
that sparsity might be a desired feature in many practical applications.
For instance, imposing sparsity on conventional EM-type learning has
been shown to produce better results part of speech tagging
applications~\cite{Vasvwani2010}. Whereas ~\cite{Vasvwani2010} had to
impose sparsity via an additional penalty term in the objective
function, in our case sparsity is a natural outcome of maximizing the
likelihood of the best sequence. While our results were obtained on a
class of exactly-solvable model, it is plausible that they hold more
generally. 

The fact that VT provides simpler
and more definite solutions|among all choices of the parameters
compatible with the observed data|can be viewed as a type of the Occam's
razor for the parameter learning. Note finally that statistical mechanics intuition behind these results is that the
aposteriori likelihood is (negative) zero-temperature free energy
of a certain physical system. Minimizing this free energy makes physical
sense: this is the premise of the second law of thermodynamics that
ensures relaxation towards a more equilibrium state.  In that
zero-temperature equilibrium state certain types of motion are frozen,
which means nullifying the corresponding transition probabilities. In
that way the second law relates to the Occam's razor. Other connections
of this type are discussed in \cite{dom}.

\subsubsection*{Acknowledgments}

This research was supported in part by the US ARO MURI grant No. W911NF0610094 and US DTRA grant HDTRA1-10-1-0086.

\newpage
\setcounter{page}{1}
\renewcommand{\thepage}{S-\arabic{page}}

\section*{Supplementary Material}

\label{zeta_section}

Here we recall how to calculate the moment-generating
function $\Lambda(n)$ via zeta-function \cite{ruelle} and periodic orbits \cite{artuso,armen}. Let $\lambda[A]$ be the
maximal eigenvalue of matrix $A$ with non-negative
elements \cite{horn}. Since $AB$ and $BA$ have identical eigenvalues, we get
$\lambda[A^d]=(\lambda[A])^d$, $\lambda[AB]=\lambda[BA]$ ($d$ is an
integer). 

Recall the content of section \ref{tro}.
Eqs.~(\ref{4}, \ref{transfer}, \ref{mubarak}) lead to
\BEA
\Lambda^m(n,m)= {\sum}_{x_1,\ldots,x_m} \phi[x_1,\ldots,x_m], \\
\phi[x_1,\ldots,x_m] \equiv \lambda\left [{\prod}_{k=1}^m T_{\x_k}\right]\lambda^n\left [{\prod}_{k=1}^m \hat{T}_{\x_k}\right]  
\EEA
where we have introduced a notation $T_{x}=T(x)$ for better readability. 
We obtain
\BEA
\phi[{\bf x'},{\bf x''}]= \phi[{\bf x''},{\bf x'}], \qquad
\phi[{\bf x'},{\bf x'}]=\phi^2[{\bf x'}],
\EEA
where ${\bf x'}$ and ${\bf x''}$ are arbitrary sequences of symbols $x_{i}$.
One can prove for $\Lambda^m(n,m)$ \cite{ruelle}:
\BEA
\label{kaban}
\Lambda^m(n,m)=\sum_{k|m}\,\,\,\sum_{(\gamma_1,...,\gamma_k) \in {\rm Per}(k) } k\left[\,
\phi[\gamma_1, \ldots, \gamma_k]\,
\right]^{\frac{m}{k}},\nonumber
\EEA
where $\gamma_i=1,...,M$ are the indices referring to realizations
of the HMM, and 
where $\sum_{k|m}$ means that the summation goes over all $k$ that
divide $m$, e.g., $k=1,2,4$ for $m=4$.  Here ${\rm Per}(k)$ contains
sequences $\Gamma=(\gamma_1,...,\gamma_k)$
selected according to the following rules: {\it i)} $\Gamma$ turns to
itself after $k$ successive cyclic permutations, but does not turn to
itself after any smaller (than $k$) number of successive cyclic
permutations; {\it ii)} if $\Gamma$ is in ${\rm Per}(k)$, then ${\rm
Per}(k)$ contains none of those $k-1$ sequences obtained from $\Gamma$
under $k-1$ successive cyclic permutations. 

Starting from (\ref{kaban}) and introducing notations $p=k$,
$q=\frac{m}{k}$, we transform $\xi(z,n)$ as 
\BEA
\xi(z,n)=\exp\left[-
\sum_{p=1}^\infty 
\sum_{ \Gamma \in {\rm Per}(p) }\, \sum_{q=1}^\infty \frac{z^{pq}}{q}
\phi^q[\gamma_1, \ldots, \gamma_p]
\right].\nonumber
\label{trevoga_1}
\EEA
The summation over $q$,
${\sum}_{q=1}^\infty \frac{z^{pq}}{q}
\phi^q[\gamma_1,\ldots, \gamma_p]
=-\ln\left[
1-z^p \phi[\gamma_1,\ldots, \gamma_p]
\right]$, yields
\begin{gather}
\xi(z,n)={\prod}_{p=1}^\infty \,
{\prod}_{\Gamma \in {\rm Per}(p) }\, 
\left[
1-z^p \phi[\gamma_1,\ldots, \gamma_p]\,
\right]\nonumber\\
\label{retmon}
=1-z{\sum}_{l=1}^M \la_l\hl_l
+{\sum}_{k=2}^\infty \epf_k z^k,
\end{gather}
where 
$\la_{\alpha...\beta}\equiv \la [T_{\x_{\alpha}}...T_{\x_{\beta}}]$,
$\la_{\alpha+\beta}\equiv \la [T_{\x_{\alpha}}]\la[T_{\x_{\beta}}]$
(all the notations introduced generalize|via introducing a hat|to functions
with trial values of the parameters, e.g., $\hat{T}_2$).
$\epf_k$ are obtained from (\ref{retmon}). We write them down assuming that $M=2$ (two realizations of the observed process)
\BEA
\label{fox}
\epf_2&=&-\la_{12}\hl_{12}+\la_{1+2}\hl_{1+2},\\
\epf_3&=&\la_{2+21}\hl_{2+21}-\la_{221}\hl_{221}
+\la_{1+12}\hl_{1+12}-\la_{112}\hl_{112},\\
\epf_4&=&-\la_{1222}\hl_{1222}+\la_{2+122}\hl_{2+122} + \la_{1+122}\hl_{1+122} - \la_{1122}\hl_{1122} \nonumber\\
&+& \la_{2+211}\hl_{2+211} -\la_{1+2+12}\hl_{1+2+12} +  \la_{1+211}\hl_{1+211} -\la_{1112}\hl_{1112}.
\label{wolf}
\EEA
The algorithm for calculating $\epf_{k\geq 5}$ is straighforward~\cite{armen}.
Eqs.~(\ref{fox}--\ref{wolf}) for $\epf_{k\geq 4}$
suffice for approximate calculation of (\ref{retmon}), where
the infinite sum ${\sum}_{k=2}^\infty$ is approximated by its first few terms.

We now calculate $\xi(z,n)$ for the specific model considered in Section~\ref{garun_al_rashid}. For this model, only the first row
of $T_1$ consists of non-zero elements, so we have 
\BEA
\lambda_{1\chi 1\sigma}=\lambda_{1\chi +1\sigma}, \qquad \hla_{1\chi 1\sigma}=\hla_{1\chi +1\sigma},
\label{korund}
\EEA
where $\chi$ and $\sigma$ are arbitrary sequences of $1$'s and $2$'s.
The origin of (\ref{korund}) is that the transfer-matrices
$T(1)T(\chi_1)T(\chi_2)\ldots$ and $T(1)T(\sigma_1)T(\sigma_2)\ldots$
that correspond to $1\chi$ and $1\sigma$, respectively, have the same
structure as $T(1)$, where only the first row differs from zero. For
$\epf_k$ in (\ref{retmon}) the feature (\ref{korund}) implies
\BEA
\epf_k &=& -\la^n [\hat{T}_1\hat{T}_2^{k-1} ]\,\la [T_1T_2^{k-1}] \nonumber \\
&+&\la^n [\hat{T}_1\hat{T}_2^{k-2} ]\,\la[T_1T_2^{k-2}]\,
\la^n[\hat{T}_2]\la[T_2].
\label{buk}
\EEA
To calculate $\la\left[T_1T_2^{p}\right]$ for an integer $p$ one diagonalizes
$T_2$ \cite{horn} 
(the eigenvalues of $T_2$ are generically not
degenerate, hence it is diagonalizable), 
\BEA
T_2={\sum}_{\alpha=1}^L
\tau_\alpha |R_\alpha\rangle\langle L_\alpha|, 
\EEA
where $\tau_\alpha$ are
the eigenvalues of $T_2$, and where $|R_\alpha\rangle$ and
$|L_\alpha\rangle$ are, respectively, the right and left eigenvectors:
\BEA
T_2|R_\alpha\rangle=\tau_\alpha|R_\alpha\rangle,\,\, \langle L_\alpha|T_2=\tau_\alpha\langle L_\alpha|,\,\,
\langle L_\alpha|R_\beta\rangle=\delta_{\alpha\beta}. \nonumber
\EEA
Here $\delta_{\alpha\beta}$ is the Kronecker delta. Note that
generically $\langle L_\alpha|L_\beta\rangle\not=\delta_{\alpha\beta}$
and $\langle R_\alpha|R_\beta\rangle\not=\delta_{\alpha\beta}$. Here
$\langle L_\alpha|$ is the transpose of $| L_\alpha\rangle$, while
$|R_\alpha\rangle\langle L_\alpha|$ is the outer product.

Now $\la\left[T_1T_2^{p}\right]$ reads from (\ref{mack2}):
\BEA
\la\left[T_1T_2^{p}\right] = {\sum}_{\alpha=1}^L \tau_\alpha^p \psi_\alpha, \quad 
\psi_\alpha \equiv \langle 1 | T_1|R_\alpha\rangle\langle L_\alpha|1\rangle ,
\label{moj}
\EEA
where $\langle 1|=(1,0,\ldots,0)$.
Combining (\ref{moj}, \ref{buk}) and (\ref{retmon}) we arrive at  (\ref{raa}).

\end{document}